# An End-to-End Two-Stream Network Based on RGB Flow and Representation Flow for Human Action Recognition

Song-Jiang Lai[1,2], Tsun-Hin Cheung[1], Ka-Chun Fung[2], Tian-Shan Liu[2], Kin-Man Lam[1,2]

[1]*Center for Advances in Reliability and Safety, New Territories, Hong Kong*

[2]*Department of Electrical and Electronic Engineering, The Hong Kong Polytechnic University. Kowloon, Hong Kong*

*Abstract*—With the rapid advancements in deep learning, computer vision tasks have seen significant improvements, making two-stream neural networks a popular focus for video-based action recognition. Traditional models using RGB and optical flow streams achieve strong performance but at a high computational cost. To address this, we introduce a representation flow algorithm to replace the optical flow branch in the egocentric action recognition model, enabling end-to-end training while reducing computational cost and prediction time. Our model, designed for egocentric action recognition, uses class activation maps (CAMs) to improve accuracy and ConvLSTM for spatio-temporal encoding with spatial attention. When evaluated on the GTEA61, EGTEA GAZE+, and HMDB datasets, our model matches the accuracy of the original model on GTEA61 and exceeds it by 0.65% and 0.84% on EGTEA GAZE+ and HMDB, respectively. Prediction runtimes are significantly reduced to 0.1881s, 0.1503s, and 0.1459s, compared to the original model's 101.6795s, 25.3799s, and 203.9958s. Ablation studies were also conducted to study the impact of different parameters on model performance.

## I. INTRODUCTION

Early action recognition techniques primarily focused on detecting and representing interest points using methods such as spatio-temporal point detection [1] and optical flow histograms [2]. Although these methods are effective for feature extraction, they struggle with large datasets due to their lack of scalability. Heng et al. [3] introduced dense trajectory features, which improve motion representation but still face scalability issues. Recent advances in deep learning have surpassed traditional methods. Karpathy et al. [4] developed deep models that process temporal information and combine dense optical flow with deep features. Despite these advancements, limitations persisted, including challenges with long videos and pre-computed optical flow. Innovations, such as spatial-temporal CNN integration [5] and LSTM-based time fusion [6], further advanced the research field of action recognition.

Zhou et al. [7] improved action recognition accuracy through weighted fusion and sequential reasoning. Temporal attention models and ConvLSTM variants [8] have enhanced the focus on important frames and preserved spatial structures, proving valuable in applications like anomaly and violence detection. Concurrently, research on optical flow has evolved, with Fan et al. [9] enhancing accuracy with the TV-L1 method, and Sun et al. [10] introducing the optical flow-guided feature (OFF). Piergiovanni et al. [11] advanced this with the representation flow algorithm, offering a faster and more accurate alternative to traditional optical flow.

Our model combines the strengths of the egocentric activity recognition model proposed by Swathikiran et al. [12] with the representation flow algorithm, yielding an end-to-end trainable system. By fusing input RGB frames with representation flow, our method not only reduces prediction time but also enhances accuracy. This is achieved by leveraging the combined benefits of both the original EgoRCNN model [12] and the advanced representation flow model.

## II. METHODOLOGY

The proposed activity recognition method comprises two components: deep feature encoding for the RGB stream employing spatial attention and class activation maps (CAMs), and a fully differentiable convolutional layer for the representation stream, which is developed from optical flow methods. The entire framework is designed to be trainable end-to-end, allowing all parameters to be optimized jointly to enhance model performance.

### A. Class Activation Maps

A novel mechanism known as Class Activation Mapping (CAM) is introduced by Zhou et al. [13] to produce the class-specific saliency maps by leveraging the average pooling layer in contemporary deep CNN architectures, such as ResNet. Let $w_l^c$ represent the weight for unit $l$ corresponding to class $c$, and $a_l(i)$ denote the activation values at spatial location $i$ of unit $l$ in the last convolutional layer. Furthermore, the CAM for class $c$, $F_l(i)$, can be calculated as follows:

$$F_c(i) = \sum_l w_l^c a_l(i). \tag{1}$$

Therefore, the CAM, $F_c$, can be used to identify the image regions that the CNN utilized to recognize the class was being considered (i.e., class c). Subsequently, we select the class that exhibits the highest probability, referred to as the winning class. This allows us to produce a saliency map using the CAM.

B. *ConvLSTM Module*

The structural block diagram of the RGB stream shows how RGB frames are encoded, is shown in **Figure 1**. ResNet-34 serves as the backbone, responsible for generating spatial attention maps, extracting relevant features, and predicting class categories. The training is conducted in two stages: Stage 1 focuses on training only the classifier and ConvLSTM layer, while Stage 2 involves training the additional convolutional and fully connection layers of the ResNet-34 network. The pre-trained weights based on the ImageNet dataset are utilized by the rest parts of the model. ConvLSTM is used for temporal aggregation, and Equation (1) is employed to compute the CAM for each RGB frame. The CAM is converted into an attention map via the softmax function along the spatial dimension, which is then multiplied with the output of the last convolutional layer. The attention map is computed as follows:

$$f_{SA}(i) = f_i \odot \frac{e^{M_c(i)}}{\sum_{i'} e^{M_c(i)}}, \tag{2}$$

Where $f_i$ represents the output feature for the final layer of ResNet-34 at the spatial location $i$. $f_{SA}(i)$ denotes the Hadamard product of $f_i$ and the softmax output, which indicates the spatial attention map applied to image features. $M_c(i)$ indicates the use of the winning category $c$ to obtain the corresponding CAM. The symbol $\odot$ is used to denote the Hadamard product. After obtaining visual features with spatial attention approximated by Equation (2), temporal encoding for frame-level information is required.

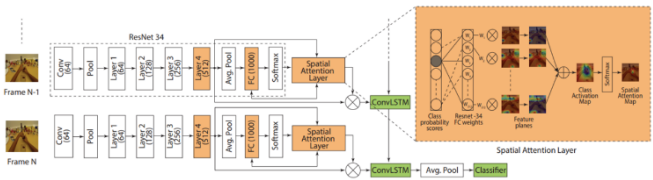

Figure 1: The RGB stream structure of the EgoRCNN model [12].

ConvLSTM works as shown by these following equations:

$$I_t = \sigma(w_x^i * f_{SA} + w_h^i * h_{t-1} + b^i), \tag{3}$$

$$f_t = \sigma(w_x^f * f_{SA} + w_h^f * h_{t-1} + b^f), \tag{4}$$

$$\tilde{c}_t = \tanh(w_x^{\tilde{c}} * f_{SA} + w_h^{\tilde{c}} * h_{t-1} + b^{\tilde{c}}), \tag{5}$$

$$C_t = \tilde{c}_t \odot f_{SA} + C_{t-1} \odot f_t, \tag{6}$$

$$O_t = \sigma(w_x^o * f_{SA} + w_h^o * h_{t-1} + b^o), \tag{7}$$

$$h_t = O_t \odot \tanh(C_t), \tag{8}$$

Where $\sigma$ denotes the sigmoid function, $I_t$, $f_t$, and $O_t$ are the input state, forget state, and hidden state, respectively, $w$ and $b$ represent the learnable weights and biases of the ConvLSTM.

C. *The Representation Flow Algorithm*

The representation flow algorithm shares many similarities but also has notable differences with the traditional optical flow algorithm. As detailed by Piergiovanni et al. [11], while both algorithms aim to capture motion information between frames, the representation flow layer is designed to extract features from deep feature maps, whereas optical flow is applied to video frames in the spatial domain. Additionally, the representation flow algorithm differs in its learning parameters: optical flow involves fixed hyper-parameters, such as $\tau$, $\theta$, and $\lambda$, which are manually set, whereas the representation flow algorithm utilizes trainable parameters that are updated during model training.

Denote $F_1$ and $F_2$ as the feature maps from two consecutive frames, with a single channel. These feature maps are then convolved with the Sobel operators, as shown in Equation (9), to compute the gradients of the feature maps.

$$\nabla F_{i,x} = \begin{bmatrix} 1 & 0 & -1 \\ 2 & 0 & -2 \\ 1 & 0 & -1 \end{bmatrix} * F_i, \quad \nabla F_{i,y} = \begin{bmatrix} 1 & 2 & 1 \\ 0 & 0 & 0 \\ -1 & -2 & -1 \end{bmatrix} * F_i. \tag{9}$$

Firstly, we introduce the flow field: $u$, and the dual vector fields: $p$, which are used to minimize the energy, and set them to zero. Then, **Algorithm 1** is applied iteratively until the minimum total variational energy is achieved.

---

**Algorithm 1: Method for the representation flow layer [11]**
Initialize $u$ and $p$, as follows:

$$u = 0, p = 0 \tag{10}$$

Compute the feature gradients using Equation (9), and then compute the difference as follows:

$$\rho_c = F_2 - F_1 \tag{11}$$

Perform the following operations, with $n$ iterations:

$$\rho = \rho_c + \nabla_x F_2 \cdot u_x + \nabla_y F_2 \cdot u_y \tag{12}$$

$$v = \begin{cases} u + \lambda\theta\nabla F_2, & \rho < -\lambda\theta|\nabla F_2|^2 \\ u - \lambda\theta\nabla F_2, & \rho > \lambda\theta|\nabla F_2|^2 \\ u - \rho\frac{\nabla F_2}{|F_2|^2}, & \text{otherwise} \end{cases} \quad (13)$$

$$u = v + \theta \cdot divergence(p) \quad (14)$$

$$p = \frac{p + \frac{\tau}{\theta}\nabla u}{1 + \frac{\tau}{\theta}|u|} \quad (15)$$

Finally, after the loop, the flow $u$ is obtained.

---

Following **Algorithm 1**, zero-pad $p$ in the $x$ and $y$ directions and convolute them with the weights $w_x$ and $w_y$ to calculate the divergence:

$$divergence(p) = p_x * w_x + p_y * w_y, \quad (16)$$

$\nabla u_x$ and $\nabla u_y$ are computed as follows:

$$\nabla u_x = \begin{bmatrix} 1 & 0 & -1 \\ 2 & 0 & -2 \\ 1 & 0 & -1 \end{bmatrix} * u_x, \nabla u_y = \begin{bmatrix} 1 & 2 & 1 \\ 0 & 0 & 0 \\ -1 & -2 & -1 \end{bmatrix} * u_y. \quad (17)$$

### D. Computing Flow-of-Flow

In principle, continuously stacking representation flow layers can enhance motion information extraction and representation, as more flow layers can more comprehensively capture the direction and magnitude of motion. However, a greater number of flow layers often results in more intricate motion features and textures. Applying the representation flow technique to sequential flow images may lead to inconsistent optical flow and non-rigid movements, hence impairing model performance. Inserting standard convolution layers between representation flow layers aids in preserving optical flow consistency and enhancing motion representation to resolve this issue. The comprehensive architecture of the video-CNN, integrated with the representation flow, which is demonstrated in **Figure 2**.

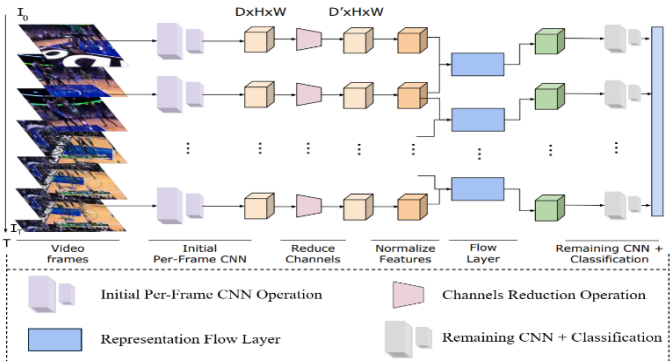

**Figure 2: The comprehensive architecture of a video recognition CNN model with a representation flow layer [11].**

### E. Applying Representation Flow Layer for Action Recognition

In our proposed model, as shown in **Figure 3**, the representation flow layer replaces the original optical flow branch in a CNN model, integrating into the EgoRCNN framework. This refined model combines the RGB and representation flow branches, employing two classifier layers for joint training. Both branches take RGB frames as input. The CNN produces predictions for each time step, which are then averaged throughout the time dimension for generating the category probabilities. The hybrid network is trained by reducing the cross-entropy loss, as outlined below:

$$L(v,c) = -\sum_{i}^{k}(c == i)\log(p_i), \quad (18)$$

Where $M$ is the classification model, $v$ represents the video, $c$ denotes a specific category among all categories, $L$ refers to the value of the cross-entropy loss function, and $p = M(v,c)$.

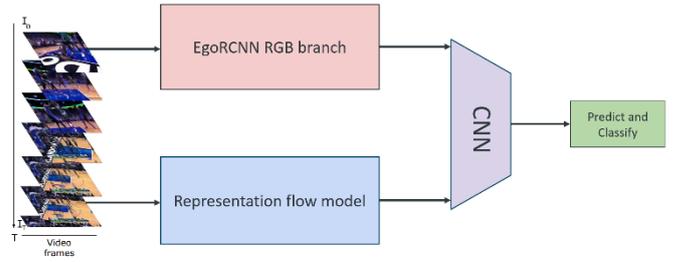

**Figure 3: The overall structure of proposed model.**

## III. EXPERIMENT RESULTS

### A. Implementation Details

In this study, we employ RGB features and motion features, based on representation flow, for action recognition. The ConvLSTM with 512 hidden units is utilized by the RGB branch of our model for spatio-temporal coding and ResNet-34 as the backbone for extracting frame-level image features and spatial attention maps. Training is performed in two stages: Stage 1 spans 200 epochs with a learning rate of $10^{-3}$, adjusted at the 25, 75, and 150 epochs, while Stage 2 involves 150 epochs with a learning rate of $10^{-4}$, adjusted at the 25 and 75 epochs, utilizing the ADAM optimizer and a batch size of 32 with 25 frames sampled per video.

The representation flow branch utilizes ResNet-34 as the backbone and consists of two representation flow layers, separated by an intermediate convolutional layer to smooth the flow. The input video clip length is set to 16 frames for optimal performance. The number of channels is reduced to 32 by a $1 \times 1$ convolutional layer, balancing speed and performance, while a $3 \times 3$ convolutional layer restores the original channel count. The flow branch is trained for 750 epochs with a batch size of 16, employing stochastic gradient descent with a learning rate of $10^{-3}$ and a decay factor of 0.5.

Then fuse the two branches at the decision-making level by adding a classifier to the top of the RGB branch and a new fully connected layer to integrate the outputs after separately training the RGB and representation flow branches. The final training phase consists of 250 epochs with a batch size of 32, utilizing a learning rate of 1 and a decay factor of 0.1 per epoch.

### B. Experimental Results

#### GTEA61

Initially, we assessed the efficacy of our model using split2 of the GTEA61 dataset and contrasted it with the original EgoRCNN model that included the optical flow branch. The findings shown in **Table 1** indicate that the representation flow model's accuracy on split2 is 37.74, while the optical flow component of the original EgoRCNN model attains 39.42. Following the collaborative training of the two streams, both our proposed model and the original EgoRCNN model attained an accuracy of 69.84. Since the size of GTEA 61 dataset is too small, the superiority of our model cannot be reflected.

| Model | Accuracy(%) | Time(s) |
|---|---|---|
| Optical flow | 39.42 | N/A |
| Representation flow | 37.74 | N/A |
| Two- stream (optical flow) | 69.84 | 101.6795 |
| Two-stream (representation flow) | 69.84 | 0.1881 |

Table 1: Comparison of the original and proposed two-stream models in terms of accuracy and inference runtime on the GTEA61 dataset

The runtime required for classifying samples demonstrates significant differences between the models. Extracting optical flow is notably time-intensive, with the original model requiring 101.55 seconds pre video for this process. As shown in Table 1, the average runtime for classifying a sample using the original model is 0.1295 seconds, resulting in a total average runtime of 101.6795 seconds when optical flow extraction is performed. In contrast, our proposed model, which eliminates the need for optical flow extraction, achieves a significantly shorter average runtime of only 0.1881 seconds.

#### EGTEA GAZE+

We trained and assessed the proposed model on the larger EGTEA GAZE+ dataset. This dataset is significantly larger than the GTEA61 dataset, providing a more comprehensive basis for our analysis. The prediction accuracies of the two models are tabulated in **Table 2**. The assessment was conducted across three splits of the EGTEA GAZE+ dataset. The results illustrate the proposed model consistently outperforms the original model in terms of accuracy.

| Method | Split1 | Split2 | Split3 | Average |
|---|---|---|---|---|
| Two-stream (optical flow) | 61.76 | 61.38 | 58.17 | 60.44 |
| Two-stream (representation flow) | 62.49 | 61.64 | 59.15 | 61.09 |

Table 2: Comparison of the original and proposed two-stream models in terms of accuracy on the EGTEA GAZE+ dataset

**Table 3** shows that the original model requires an average runtime of 0.0999 seconds to predict the class category of a sample. However, it necessitates an additional 25.28 seconds for the extraction of optical flow, bringing the total runtime to 25.3799 seconds. In contrast, our proposed model exhibits remarkable efficiency, requiring only 0.1503 seconds for the classification process. This also represents a significant improvement in terms of speed and efficiency.

| Method | Time(s) |
|---|---|
| Two-stream (joint train with optical flow) | 25.3799 |
| Two-stream (joint train with representation flow) | 0.1503 |

Table 3: Comparison of the original and proposed two-stream models in terms of prediction runtime on the EGTEA GAZE+ dataset

#### HMDB

The superiority of our proposed model is previously assessed using the GTEA61 and EGTEA GAZE+ datasets, both of which are first-person perspective datasets. To further illustrate the generalizability and robustness of our proposed model, we expanded our experiments to the HMDB dataset, which provides a third-person perspective and has a modest sample size. The experimental findings on the HMDB dataset are shown in **Table 4**, illustrating the performance of our proposed model across various perspectives and sample sizes.

Our proposed model outperforms the original model can also be demonstrated by the prediction results on the HMDB dataset. Specifically, while the original model achieved an accuracy of 22.89, our proposed model reached 24.34. With joint training of both two streams, the accuracy for the original model improves to 49.87, compared to 50.71 for the proposed model. In terms of runtime, as shown in **Table 4**, the original model takes an average runtime of 203.9958 seconds, including 0.1058 seconds for classification and 203.89 seconds for optical flow extraction. In contrast, the proposed model significantly reduces the average runtime to only 0.1459 seconds.

| Methods | Accuracy (%) | Time (s) |
|---|---|---|
| RGB flow only | 50.00 | N/A |
| Optical flow only | 22.89 | N/A |
| Representation flow only | 24.34 | N/A |
| Two-stream with optical flow | 49.87 | 203.9958 |
| Two-stream with representation flow | 50.71 | 0.1459 |

Table 4: Comparison of the original and proposed two-stream models in terms of accuracy and runtime on the HMDB dataset

### C. Ablation study

**Optimal number of representation flow layers**

The number of representation flow layers incorporated into a CNN model significantly impacts its performance. We investigated the effects of setting different number of representation flow layers to 1, 2, 3 and none on our proposed model. Evaluations were conducted on the HMDB dataset using consistent experimental settings, and the results are tabulated in **Table 5**. Our findings indicate the model with two representation flow layers achieved the highest accuracy compared to the configurations with 1, 3 or no flow layers.

| Configuration | Accuracy(%) |
|---|---|
| Without representation flow layer | 20.08 |
| 1 representation flow layer | 23.15 |
| 2 representation flow layers | 24.34 |
| 3 representation flow layers | 23.11 |

Table 5: Recognition accuracy of the proposed model with different numbers of representation flow layers on the HMDB dataset

**Optimal number of iterations for training**

The performance of our model highly relies on the number of iterations used to compute the flow in the representation flow layer. We conducted experiments using 10, 20, 30, and 50 iterations, and the results are shown in **Table 6**. It is clear our model achieves the highest accuracy when 20 iterations are adopted for computing representation flow.

| Number of iterations | Accuracy (%) |
|---|---|
| 10 | 21.58 |
| 20 | 24.34 |
| 30 | 23.78 |
| 50 | 23.14 |

Table 6: Recognition accuracy of the proposed model with different number of iterations used for computing representation flow on the HMDB dataset

**Effects of backbone size**

The depth of a deep model can influence its tendency to overfit, thus affecting its accuracy and generalization. In order to investigate this, we implemented experiments that involved altering the size of backbone for the representation flow branch from ResNet-18 to ResNet-101. The performance is influenced by the model depth, as demonstrated in **Table 7**. It is evident the accuracy of the proposed model is at its highest when the backbone of our model is configured to ResNet-34.

| Backbone size | Accuracy (%) |
|---|---|
| ResNet-18 | 21.03 |
| ResNet-34 | 24.34 |
| ResNet-50 | 24.27 |
| ResNet-101 | 22.89 |

Table 7: Recognition accuracy of the proposed model with backbone of different sizes on the HMDB dataset

**Comparison to other state-of-art methods**

**Table 8** and **Table 9** summarize the results of numerous experiments conducted to assess the efficacy of our proposed model and compare it to state-of-the-art methods. The methods in **Table 8** focus on egocentric activity recognition tasks, while **Table 9** addresses the third-person activity recognition tasks. The experiment results in Table 8 illustrate the accuracies on the fixed split2 of GTEA61, and for **Table 9**, it shows the prediction accuracies on each split of EGTEA GAZE+.

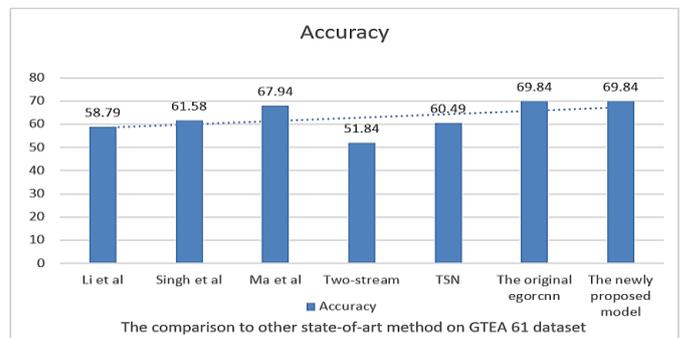

Table 8: Comparison of the proposed model with state-of-art methods in terms of prediction accuracy (%) on the GTEA 61 dataset

Li et al. [14] proposed a technique that uses hand segmentation and gaze information for detecting object locations in the first two datasets, alongside the Gaze+ dataset. Ma et al. [15] developed a network that integrates object localization and hand segmentation to enhance the accuracy of object and self-location detection. In contrast, Singh et al. [16] introduced the ego-flow input into a two-stream model for egocentric motion recognition tasks. Besides, Carreira et al. [17] introduced the I3D method, which extends 2D convolutions to 3D, enabling spatio-temporal feature learning for improved video action recognition. The TSN structure sparsely samples video clips and aggregates their predictions for video-level human action recognition is proposed by Limin Wang et al. [18]. In the prediction accuracy results from **Table 8** and **Table 9**, we can observe a significant improvement when comparing the proposed model to other state-of-the-art methods.

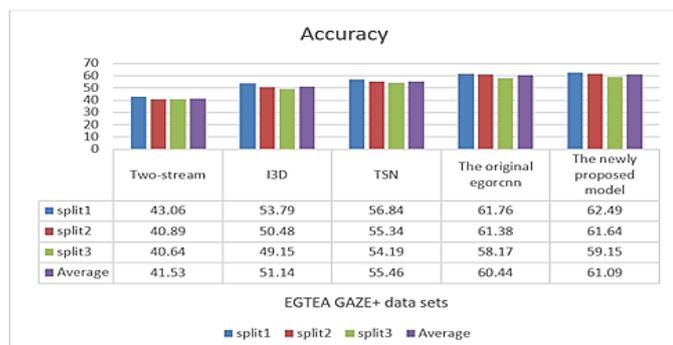

Table 9: Comparison of accuracy of the proposed model and state-of-the-art methods across different splits on the EGTEA GAZE+ data sets

## IV. CONCLUSIONS

In this study, we have proposed a learnable representation flow layer as a replacement for the traditional optical flow branch in the original EgoRCNN model. Our model leverages the class attention map (CAM) of the RGB stream, thereby enhancing its ability to focus on activity-relevant regions. We further incorporate Conv-LSTM for spatio-temporal encoding with spatial attention. This has resulted in an improved performance over the original EgoRCNN model in terms of both accuracy and prediction runtime. This improvement is evident across three standard datasets: two of which are from the first-person perspective and the remaining one from the third-person perspective.